\pdfoutput=1

\documentclass[11pt]{article}

\usepackage[final]{coling}

\usepackage{times}
\usepackage{latexsym}
\usepackage{tipa,vowel}
\usepackage{hyperref}
\usepackage[T1]{fontenc}

\usepackage[utf8]{inputenc}

\usepackage{microtype}

\usepackage{inconsolata}

\usepackage{graphicx}

\title{Dialectal and Low-Resource Machine Translation for Aromanian}

\usepackage{adjustbox} 
\usepackage[normalem]{ulem}
\useunder{\uline}{\ul}{}
\usepackage{lscape}
\usepackage{tabularx}
\usepackage{amsmath}
\usepackage{multirow}

\author{
 \textbf{Alexandru-Iulius Jerpelea\textsuperscript{$\ddagger$}}
 \textbf{Alina Rădoi\textsuperscript{$\mathsection$}} 
 \textbf{Sergiu Nisioi\textsuperscript{$\star$}}\thanks{Corresponding author.}\\
 \\
 \textsuperscript{$\ddagger$} Tudor Vianu National College of Computer Science, Bucharest \\
 \textsuperscript{$\mathsection$} West University of Timișoara \\
 \textsuperscript{$\star$} Human Language Technologies Research Center, University of Bucharest \\
 \texttt{\normalsize{alex.jerpelea@gmail.com}}, \texttt{\normalsize{alina.radoi@e-uvt.ro}}, \texttt{\normalsize{sergiu.nisioi@unibuc.ro}}
}

\begin{document}
\maketitle

\begin{abstract}
This paper presents the process of building a neural machine translation system with support for English, Romanian, and  Aromanian - an endangered Eastern Romance language. 
The primary contribution of this research is twofold: (1) the creation of the most extensive Aromanian-Romanian parallel corpus to date, consisting of 79,000 sentence pairs, and (2) the development and comparative analysis of several machine translation models optimized for Aromanian.
To accomplish this, we introduce a suite of auxiliary tools, including a language-agnostic sentence embedding model for text mining and automated evaluation, complemented by a diacritics conversion system for different writing standards. This research brings contributions to both computational linguistics and language preservation efforts by establishing essential resources for a historically under-resourced language. All datasets, trained models, and associated tools are public:\footnote{\url{https://huggingface.co/aronlp}} \url{https://arotranslate.com}
\end{abstract}

\section{Introduction}

Training good machine translation (MT) systems in a low-resource setting is a task far from being solved \cite{ijcai2021p629,haddow-etal-2022-survey}. Current advances in Large Language Models (LLMs) and evaluation methodologies are centred on English or are massively multilingual, but do not engage with the particularities of low-resource languages. Of the 7000+ languages spoken in the world, only a small part is covered by current MT systems \cite{costa2022no}. 

Aromanian (ISO 639-3 - rup) is an endangered Eastern Romance language \cite{moseley2010atlas}, which currently lacks large-scale corpora and electronic resources that can potentially contribute to the preservation of its cultural heritage. 
In this study, we build a dataset suitable for training MT systems in the low-resource setting between two related languages from the same dialect continuum: Romanian (ron) and Aromanian (ISO 639-3 - rup). Given the close similarity \cite{caragiucompendiu} between the two languages, we expand the existing pre-trained LLMs and machine translation models available for standard Romanian to its Eastern Romance sibling, in order to contribute to the preservation of this endangered language.\footnote{\url{https://endangeredlanguages.com/lang/963}, Accessed: 2024-09-16}

The difficulties for such a task stem from the fact that: (1) there is little availability of monolingual or bilingual texts, (2) the writing system, vocabulary, and grammar have not been standardized in a widely accepted institutional manner, several types of spelling are currently being used \cite{caragiuDictionar,cuniaScriere,nevaci2008sisteme}, and (3) Aromanian has several varieties / dialects that have been influenced by contact with Greek, Romanian, Turkish, Albanian, and South Slavic languages \cite{caragiucompendiu,pascaru_kahl_rramanj_2017}.  
Furthermore, the language has historically been transmitted orally within families or small communities \cite{gica2009recent,maiden2016romanian}, with public usage limited to a few towns in Albania and North Macedonia. 

Our contribution is: 
\begin{itemize}
  \item the creation of a 79k multigenre Aromanian-Romanian sentence-aligned parallel corpus, augmented with machine-translated English sentences;
  \item a language-agnostic BERT Sentence Encoder (LaBSE) sentence encoder \cite{feng2020language}, fine-tuned for Aromanian support;
  \item a comparison of different No Language Left Behind (NLLB) \cite{costa2022no} models fine-tuned to translate in any direction between Aromanian, Romanian, and English;
  \item a range of instruction-tuned large language models, fine-tuned for Aromanian translation; 
  \item a diacritics converter between two major orthographic variants of Aromanian.
\end{itemize}


\section{Related Work}
\label{sec:PrevWork}
Low-resource machine translation has gained considerable traction in the past few years. Initiatives range from commercial ambitions to build ``machine translation systems for the next thousand languages'' \cite{bapna2022building}, the NLLB family of models that can translate between more than 200 languages \cite{costa2022no}, to approaches focusing on severely under-resourced languages \cite{parida2021open,sanchez-martinez-etal-2024-findings}.


Varieties and closely related languages also receive special attention, from early work on dialectal and language varieties translations exploring statistical machine translation and rule-based systems \cite{zhang1998dialect,scannell2006machine,otte2011rapid,hassani2017kurdish} to more recent WMT Shared Tasks \cite{akhbardeh-etal-2021-findings} covering translations from the same language families, such as Dravidian, Manding, and Romance languages. Although rule-based methods are still a strong baseline for certain language pairs \cite{sanchez-martinez-etal-2024-findings}, neural approaches provide the state of the art in multiple tasks from Portuguese \cite{costa-jussa-etal-2018-neural}, Serbo-Croatian \cite{popovic-etal-2020-neural}, Belorussian-Russian, to Arabic dialects \cite{DBLP:journals/mt/KarakantaDG18,kumar-etal-2021-machine}.

Aromanian remains an unexplored language in MT research. The electronic resources available for Aromanian consist primarily of multilingual word-aligned lists \citep{Nisioi14,ciobanuDinu,beniamine2020opening,mititelu2021instrument,fourrier2022probing}, which are used to study its evolution, history, contact relationship with other Romance languages. The recent work of \citet{petrariu2024multilingual} presents a multilingual parallel corpus of approximately 3k sentences covering Aromanian. While it is the largest published resource to date, this corpus only covers one genre, namely fairy tales, and the total number of sentence pairs renders it insufficient for training qualitative MT models, as concluded by \citet{petrariu2024multilingual}.

To the best of our knowledge, no successful attempts have been made to integrate Aromanian into a translation system, and our work lays the first building blocks in this direction.

\section{Collecting a Multigenre Dataset}
\label{sec:sources}

The original texts collected in our dataset focus on the Romanian-Aromanian language pair and pertain to different genres such as news articles, literature, dictionaries, religious texts, music lyrics, and essays. 
Throughout the data collection process, we use a sentence embedder compatible with Aromanian to align sentences across languages. We fine-tune a language-agnostic BERT Sentence Encoder \cite{feng2020language} using a similar approach to \citet{dale2022first} (see \autoref{sec:Align}).

%

\textbf{The Bible} has been translated by Dina Cuvata and was published in 2004. We use an online scan of the printed edition published by the Aromanian Library hosted by Dini Trandu.\footnote{\url{https://dinitrandu.com/wp-content/uploads/2022/04/Bibliea-limba-armaneasca.pdf}, Accessed: 2024-09-16} 
Each page of the scan is digitized using Tesseract V5 \cite{smith2007overview}. The engine does not support Aromanian, so the language parameter is set to Romanian. Although the OCR quality is mediocre, it is high enough for us to post-process the results. We then carefully check each verse and pair it with its match from the Romanian translated Bible. Each verse is sentence-split using regular expressions in both languages. If the number of sentences per verse matches, we pair the sentences and add them to our corpus. We obtain around $30.5$k sentence pairs from this source.

\textbf{The Divine Comedy} has been translated by Dina Cuvata from Romanian.\footnote{\url{https://dinitrandu.com/wp-content/uploads/2018/08/Dante-Dina-Cuvata.pdf}, Accessed: 2024-09-16} By aligning it with the Romanian version, using multilingual sentence embeddings described in \autoref{sec:Align}, we obtain $2.2$k pairs.

\textbf{Tao Te Ching} has been partially translated from English by Mihali Prefti and was published by the Aromanian Library hosted by Dini Trandu.\footnote{\url{https://dinitrandu.com/wp-content/uploads/2022/06/Cartea-a-Calillei.pdf}, Accessed: 2024-09-16} Using automatic sentence alignment based on LaBSE (see \autoref{sec:Align}), we obtain 260 sentences.

\textbf{Lyrics Translate}\footnote{\url{https://lyricstranslate.com}, Accessed: 2024-09-16} is a platform where songs are translated into multiple languages. We scrape the 500+ songs in both Aromanian and Romanian and pair them verse by verse, obtaining $8.5$k pairs.

\textbf{The Multilingual Parallel Corpus of Aromanian (MPC-rup)} consists of approximately $3$k pairs of sentences from Aromanian fairy tales and short prose texts \cite{petrariu2024multilingual}.

\textbf{A list of parallel common phrases and idioms}, consisting of $2.1$k pairs, provided for this project by members of the Aromanian community.

\textbf{The Avdhela Project}\footnote{\url{https://www.proiectavdela.ro} Accessed: 2024-09-16} is a digital Aromanian library consisting of a collection of parallel poetry and prose texts. We use only the poems and pair them verse by verse, thus obtaining another $4$k pairs for our corpus.

\textbf{Radio Romania International} is a Romanian public news radio station that includes editorials in many languages, among them Aromanian. Most radio broadcasts are published in text form on the official website\footnote{\url{https://www.rri.ro/ro_ar}, Accessed: 2024-09-16} and Aromanian articles are usually translations of matching Romanian articles.\footnote{A fact confirmed by several authors working for the radio station.} However, there are no backlinks to these pages, making it difficult to trace the corresponding articles. Thus, articles in both languages are scraped and matched by the images they contain. In cases where multiple matches are possible, we use our sentence alignment tool (described in \autoref{sec:Align}) to align the titles in Aromanian with those in Romanian, pairing the titles with the greatest semantic similarity while allowing for unmatched titles. Then, for the sentence alignment of each pair of articles, we once again deploy the aforementioned alignment tool. This method yields $18.7$k sentence pairs.

\textbf{The Adventures of Tom Sawyer} has been translated from Romanian and the digital version was donated for this project. A total of $3.1$k sentences are obtained after automatic alignment.

\begin{figure}[t]
  \includegraphics[width=\columnwidth]{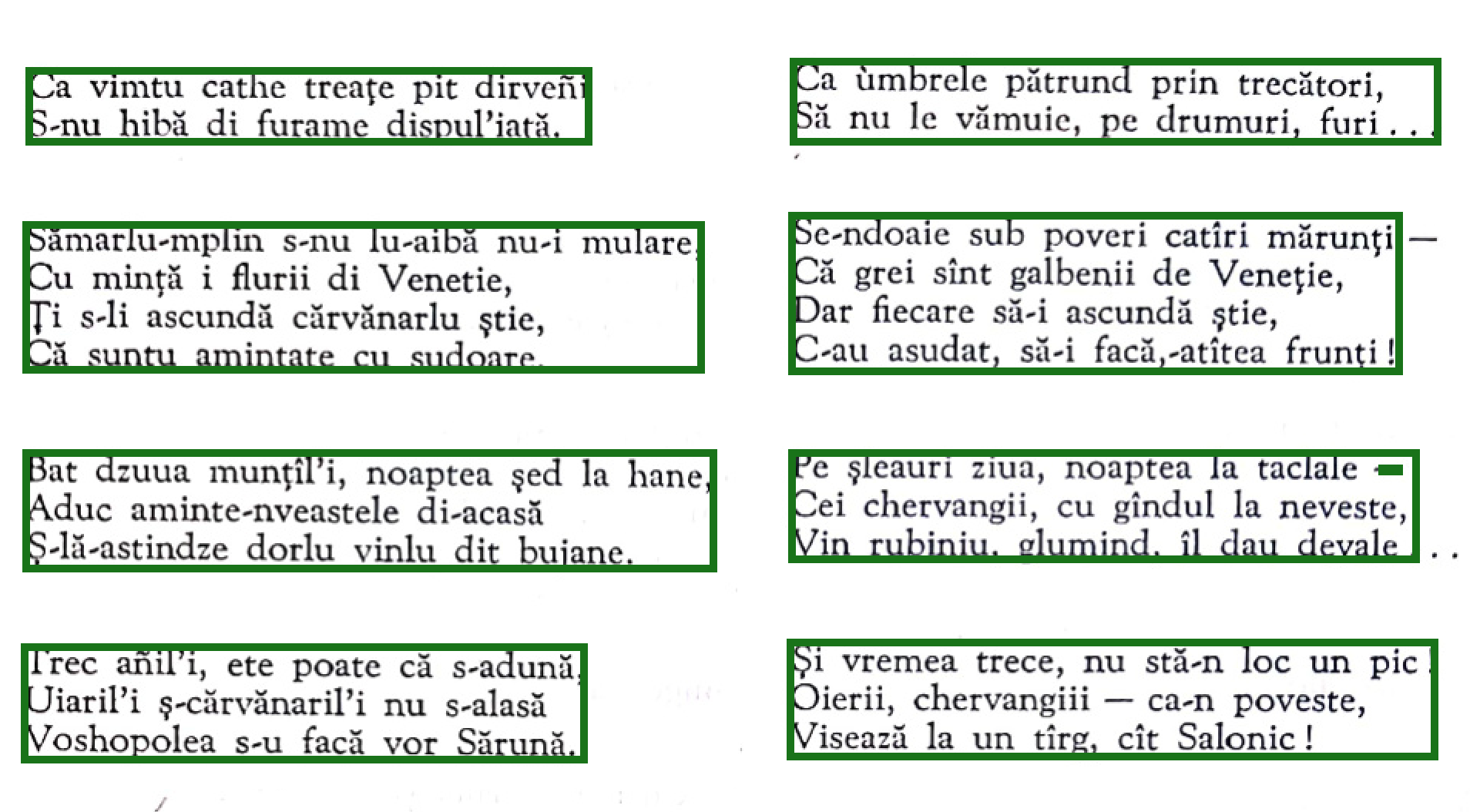}
  \caption{Detecting text bounding boxes in ``A century of Aromanian poetry''.}
  \label{fig:bounding_box}
\end{figure}

\textbf{A century of Aromanian poetry} \cite{candroveanu1985veac} is a collection of Aromanian poetry translated into Romanian. 
We use a scanned digitized version and apply Open CV \cite{opencv_library} to identify text bounding boxes (\autoref{fig:bounding_box}). We then apply OCR and the same post-processing steps used for the Bible. Finally, we align the resulting sentences for each page using the alignment tool described in \autoref{sec:Align}, adding approximately $1.7$k pairs to the corpus.

\textbf{A collection of modern Aromanian poetry} written by George Vrana
was donated for this project with the author's agreement. Aligning the text verse by verse resulted in $2.1$k pairs.

\begin{table*}
  \centering
  \resizebox{0.9\linewidth}{!}{
  \begin{tabular}{lccccccccc}
    \hline
    \textbf{Split} & \textbf{Bible} & \textbf{Lyrics} & \textbf{MPC-rup} & \textbf{Phrases} & \textbf{Avdhela} & \textbf{Radio} & \textbf{Poetry} & \textbf{Tom Sawyer} \\
    \hline
    \textit{train} & $29004$ & $8142$ & $3208$ & $2038$ & $3852$ & $17867$ & $1663$ & $2963$\\
    \textit{dev}   & $1531$ & $431$ & $169$ & $116$ & $182$ & $880$ & $102$ & $171$\\
    \textit{test}  & - & - & - & - & - & - & - & - \\\hline
  \end{tabular}
  }
  \resizebox{0.95\linewidth}{!}{
  \begin{tabular}{lccccccc}
    \hline
    \textbf{Split} & \textbf{Divine Comedy} & \textbf{Modern Poetry} & \textbf{Tao Te Ching} & \textbf{The Little Prince} & \textbf{Writings} & \textbf{Total}\\
    \hline
    \textit{train} & $2133$ & $2137$ & $246$ & - & - & $73253$ \\
    \textit{dev}   & $110$ & $115$ & $17$ & $623$ & - & $4331$ \\
    \textit{test}  & - & - & - & - & $1397$ & $1397$ \\\hline
  \end{tabular}
  }%
  \caption{\label{citation-guide}
    Number of sentence pairs for each source.  The \textit{dev} set contains a random subsample of all the genres and texts from ``The Little Prince''. The \textit{test} set contains out-of-domain mixed-genre texts.
  }
  \label{tab:distrib}
\end{table*}

\textbf{The Little Prince} has been translated into Aromanian by Maria Bara and Thede Kahl. We extract $620$ sentence pairs using automatic alignment.

\textbf{Writings} is a collection of texts consisting of news articles, personal stories, and essays. These are not available online and were donated for this project by Kira Mantsu, a prominent Aromanian writer. The $16$ documents are originally in Aromanian and translated into Romanian by the author herself; the automatic alignment results in a total of $1.9$k new pairs.

\subsection{Dataset Split}
Considering all the sources mentioned above, the dataset we use to build our models consists of $79$k sentence pairs. A table with sample sentences from each source is available in \autoref{sec:appendixA}.

In preparation for the model training process, we split our dataset into \textit{train}, \textit{dev} and \textit{test} splits. 
Sentences from ``The Little Prince'' and ``Writings'' are only included in the \textit{dev} and \textit{test} sets, but are not used for training. The reason behind this choice is that we want to evaluate the model's performance on novel data, and we believe that the literary genre is a difficult evaluation task. 

For the \textit{train} set, we extract in a stratified manner $95\%$ random pairs from each of the above sources. The remaining $5\%$ pairs from these sources make up the \textit{dev} set. We also add the sentences from ``The Little Prince'' to the \textit{dev} set to provide a difficult evaluation set of literary translations.
The \textit{test} set comprises only of the mixed-genre ``Writings'' documents to enforce evaluation on out-of-domain data; thus, the texts do not have a common authorial origin with the data on which our models are trained. \autoref{tab:distrib} contains the exact number of sentence pairs extracted from each source.

\subsection{Synthetic Data}

To perform translations between Aromanian and English, we add synthetic English counterparts to all sentence pairs by translating Romanian sentences with the Google Translate API. 

To evaluate the quality of automatic English translations, we employ cometkiwi-da model \cite{comet} for reference-free evaluations; therefore, English translations are evaluated only with regard to the original Romanian sentences. Throughout the corpus, we obtain a \textit{system-score} of $0.765$ (scores range from $0$ to $1$). The resulted score is an indicator that Romanian-to-English machine translations are of mediocre quality and that automatic translations to and from English are error-prone but still meaningful.


\subsection{Orthography}
It is important to note that there are numerous orthographic standards for Aromanian. The main spelling types in our corpus are \textit{DIARO} (named after the Aromanian-Romanian Dictionary by \citet{caragiuDictionar}, and \textit{Cunia}, named after the author of the \citet{cuniaDict} Dictionary. We also sparsely found Greek orthography, which we standardized using the Latin script.

The \textit{Cunia} and \textit{DIARO} spellings are easily convertible to each other using simple regular expressions, with the exception
of the close central and mid central vowels. Both of these sounds are represented by the grapheme <ã> (U+00E3) within the \textit{Cunia} standard. However, the \textit{DIARO} standard follows the Romanian standardization \cite{petrariu2024multilingual}.\footnote{The Romanian standardization remains inconsistently adopted, particularly in online texts. The primary variation is the alternation between <â> (used mid-word) and <î> (used at the beginning and end of words) to represent the close central vowel. The introduction of the grapheme <â> by the Romanian Academy in 1993 as an \textit{anticommunist} measure has been widely criticized by linguists for its lack of scientific and etymological justification \cite{dumistracel1993lupta}. Consequently, certain Romanian publications and publishers allow their authors to choose their preferred standard rather than enforcing this rule.
} using two different symbols <î> (U+00EE) and <â> (U+00E2) for close central and <ă> (U+0103) for mid central.

We release the corpus in both writing standards, but for training purposes the \textit{Cunia} writing system is used. The main reason for this choice is the slightly lower $2.36$ fertility rate of the tokenizers  versus the \textit{DIARO} orthography that reaches $2.52$ tokens per word. We observed lower fertility scores across various models, including both NLLB-200 and pre-trained large language models. Since the pre-trained models are multilingual, it is likely that the tokenizers have seen byte-pairs similar to the \textit{Cunia} orthography in its pretraining data, i.e., words from Albanian and romanised transliterations from South Slavic and Greek languages.

Combating the potential loss of details due to the merging of the mid central and close central vowels, we train an n-gram-based model to convert from \textit{Cunia} to \textit{DIARO}. 
\autoref{tab:stats} presents several statistics regarding our \textit{Cunia}-converted datasets. Original Aromanian texts have the greatest lexical richness and shorter phrases. However, the Romanian human translations and the machine- translated texts in English have both lower lexical diversity (as estimated by the type-token ratio) and larger sentence lengths. In \autoref{sec:appendixB}, we also present these statistics for each text genre separately.

\begin{table*}
  \centering
  \begin{tabular}{llrrr}
    \hline
    language & words & unique words & type-token ratio & words/sentence \\
    \hline
    Aromanian & 1,189,000 & 175,000 & 0.15 & 15.31 \\
    Romanian & 1,279,000 & 114,000 & 0.09  & 16.45 \\
    English & 1,371,000 & 68,000 & 0.05 & 17.64 \\
    \end{tabular}

  \caption{\label{citation-guide}
       Dataset statistics of texts converted to the \textit{Cunia} standard. Words are extracted using a regex tokenizer.
  }
  \label{tab:stats}
\end{table*}

\section{Aromanian Sentence Embeddings}
\label{sec:Align}
To embed Aromanian sentences in the same latent space as Romanian and English sentences a language-agnostic BERT Sentence Encoder \cite{feng2020language} is fine-tuned following the methodology described by \citet{dale2022first}.

First, the tokenizer's vocabulary is trained on a monolingual Aromanian corpus, using all Aromanian sentences prior to alignment. This is performed with the same WordPiece tokenizer from BERT \cite{devlin-etal-2019-bert}, yielding 4,400 new tokens. As a result, the tokens-per-word ratio of the LaBSE tokenizer decreases from $2.36$ to $1.77$.

Secondly, the model is fine-tuned on both Aromanian-Romanian and Aromanian-synthetic English parallel pairs, updating only the pooled output corresponding to Aromanian embeddings using contrastive loss.

\begin{figure}[t]
  \includegraphics[width=\columnwidth]{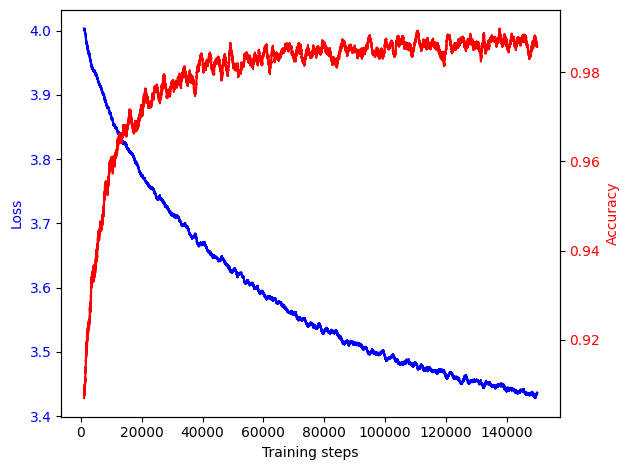}
  \caption{Average loss and accuracy during the LaBSE fine-tuning.}
  \label{fig:labseLoss}
\end{figure}

At each training step, a batch of sentence pairs is randomly selected from one of the two pairs of languages. The dot product of the embeddings for all possible sentence pairs in the batch is computed, rewarding only the pairs that are correct matches. To prevent overfitting, a small margin of $0.3$ is subtracted from the reward for matching pairs. Training stops after $150$k steps, as both the loss and accuracy graphs (for matching translated sentences) flatten beyond this point (\autoref{fig:labseLoss}). In a batch of random sentence pairs, the model pairs each Aromanian sentence with its corresponding translation with an accuracy of over $98\%$. Details on the training hyper-parameters are provided in \autoref{sec:appendixLABSE}.


The model is used in two ways: (1) to calculate the BERTScore \cite{zhang2019bertscore} and evaluate trained machine translation models (see \autoref{sec:aEval}); and (2) to mine and align sentences in parallel documents. Similarly to \citet{dale2022first}, dynamic programming is used to select a sequence of sentence pairs that have the highest possible sum of similarity scores. 

\section{Machine Translation Models}
\label{sec:experiments}

\subsection{GPT Baseline}
Inspired by recent MT results \cite{hendy2023good,kocmi2024preliminary} where closed-source systems achieved state-of-the-art performance on machine translation, we employ the GPT-4o model, in hopes of leveraging its extensive multilingual pre-training \cite{achiam2023gpt}. Additionally, it is likely that the model has been exposed to Aromanian in its pre-training stage. 
We use OpenAI's \textit{gpt-4o} model in zero-shot mode to translate all pairs in the \textit{test} split, which we later evaluate in \autoref{sec:eval}.
\begin{table*}
\centering
\resizebox{0.95\linewidth}{!}{
\begin{tabular}{p{0.2\textwidth}p{0.75\textwidth}}
    \hline
    \textbf{System prompt} & You are a helpful assistant capable of accurately translating between Aromanian, Romanian, and English. \\
    \hline
    \textbf{Instruction} & Translate this sentence from Aromanian to English \\
    \hline
    \textbf{Input} & Shi-lji intrã Enoh tu vrearea-al Dumnidzã sh-ma multu nu s-aflã cã-l mutã Dumnidza. \\
    \hline
    \textbf{Output} & And God pleased Enoch, and then he was no more, because God moved him. \\
    \hline
\end{tabular} %
}
\caption{Prompt format example for finetuning the LLaMA 3.1 Instruct model.}
\label{tab:prompt}
\end{table*}

\begin{table*}[h]
    \centering
    \begin{tabular}{lrr|rr|rr}
    \hline
    model & ron$\rightarrow$rup & rup$\rightarrow$ron & eng$\rightarrow$rup & rup$\rightarrow$eng & eng$\rightarrow$ron & ron$\rightarrow$eng \\
    \hline
    GPT-4o & 19.0 & 37.2 & 17.6 & 36.7 & \textbf{65.7} & \textbf{79.2} \\
Qwen2 7B Instruct     & 45.8                  & 50.5                  & 38.8                  & 51.9                  & 58.5                  & 76.2                  \\
TowerInstruct v0.2    & 45.7                  & 50.9                  & 31.2                  & 44.6                  & 50.8                  & 75.2                  \\
LLaMA 3.1 8B Instruct & 46.3                  & 51.4                  & 38.8                  & 52.6                  & 60.8                  & 77.2                  \\
RoLLaMA               & 46.0                  & 50.6                  & 39.7                  & 52.2                  & 60.2                  & 77.3                  \\
\hline
NLLB 1.3B             & 47.8                  & \textbf{53.2}                  & 43.6                  & \textbf{54.7}                  & 63.2                  & 76.3                  \\
NLLB 600M             & \textbf{49.1}                  & 52.7                  & \textbf{44.5}                  & 53.8                  & 62.2                  & 75.5      \\
    \hline
    \end{tabular}
    \caption{
    ChrF++ evaluation scores of different models on the \textit{test} set. Zero-shot GPT-4o is unable to produce translations into Aromanian.
    NLLB models consistently outperform LLMs on translations from and to Aromanian. No significant differences can be observed between RoLLaMA and other models that have not been previously trained on Romanian instruction data. GPT-4o achieves the best scores on Romanian and English, although the English sentences are machine translated from Romanian.
    }
    \label{chrfres_paper}
\end{table*}

\subsection{NLLB Machine Translation}

Seeking to benefit from transfer learning \cite{khiu2024predicting}, we fine-tune NLLB-200 \cite{costa2022no}, an encoder-decoder transformer architecture model that can translate between any of its 202 languages. The goal is to obtain a model that has the ability to translate between Aromanian, Romanian, and English. The last two are already supported by the NLLB.

The NLLB tokenizer uses language tags, i.e. special tokens are added to the source and target texts, which are employed by the model in the pre-training phase to identify the source and target languages. The special tokens for Romanian and English are \textit{\textless ron\_Latn\textgreater} and \textit{\textless eng\_Latn\textgreater}, respectively. We expand the tokenizer vocabulary with the \textit{\textless rup\_Latn\textgreater} token for Aromanian. Its meaning is supposedly constructed during the fine-tuning step. The embedding value of the newly added token is set to the mean of the \textit{\textless ron\_Latn\textgreater} and \textit{\textless ell\_Grek\textgreater} (Greek) embedding vectors. This is justified by the close relationship of Aromanian with standard Romanian and the large proportion of Greek influence and loanwords \cite{pascaru_kahl_rramanj_2017,BalkanRomance}.


We experiment with the 1.3B and 600M distilled models of the original Mixture-of-Experts 54B billion parameter model. Each model is trained for $100$k steps, where each sentence pair in the batch is in a random direction between Aromanian, Romanian, and English (i.e. six possible directions). At every $10$k steps, we save a checkpoint and evaluate the \textit{dev} set. The NLLB-600M checkpoint at $90$k training steps, and the NLLB-1.3B checkpoint at $70$k training steps produce the highest evaluation scores on the \textit{dev} split. The hyper-parameters used for training the two models are available in \autoref{appendixD}.

\subsection{LLMs for Machine Translation}
LLMs have proven to be the state of the art for general machine translation, from zero-shot or few-shot translation \cite{kocmi2024preliminary} to different fine-tuning strategies\cite{alves2024tower}. 

We perform full fine-tuning on four trained for instruction following:
\begin{itemize}
\setlength\itemsep{0.1em}
    \item LLaMA 3.1 8B Instruct \cite{dubey2024llama} – a multilingual model from Meta AI in which Romanian is incidentally covered in the $15\%$ of multilingual tokens used during pre-training
    \item Qwen2 7B Instruct \cite{yang2024qwen2} - a multilingual model from Alibaba Cloud covering as many as 30 languages, including 4 well-resourced Western and Italo-Romance languages: French, Spanish, Portuguese, Italian
    \item RoLLaMA 3 8B Instruct \cite{masala2024vorbe} – a LLaMA3-based model fine-tuned specifically to respond to tasks and instructions in Romanian, hoping to facilitate knowledge transfer to Aromanian
    \item TowerInstruct 7B v0.2 \cite{alves2024tower} – a LLaMA2-based model trained specifically to solve translation tasks (document-level translation, terminology-level translation, etc.) on 10 languages, including 4 well-resourced Western and Italo-Romance languages: French, Spanish, Portuguese, Italian 
\end{itemize}

\begin{table*}[h]
\centering
\begin{tabular}{l|l|ll|ll|ll}
\hline
\multirow{2}{*}{Participant}   & \multirow{2}{*}{Direction} & \multicolumn{2}{c|}{Fluency}     & \multicolumn{2}{c|}{Style}       & \multicolumn{2}{c}{Meaning}     \\ \cline{3-8} 
                           &                            & HT   & MT   & HT   & MT   & HT   & MT   \\ \hline
\multirow{2}{*}{Subject 1} & ron$\rightarrow$rup      & 9.66 & 9.73 & 9.60 & 9.60 & 9.73 & 9.86 \\ \cline{2-8} 
                           & rup$\rightarrow$ron      & 9.46 & 9.66 & 9.46 & 9.53 & 9.46 & 9.33 \\ \hline
\multirow{2}{*}{Subject 2} & ron$\rightarrow$rup      & 9.66 & 9.66 & 9.33 & 9.60 & 9.66 & 9.73 \\ \cline{2-8} 
                           & rup$\rightarrow$ron      & 8.13 & 9.0  & 7.73 & 8.73 & 8.2  & 8.93 \\ \hline
\end{tabular}
\caption{Average direct assessment scores for machine translated texts (MT) and human translations (HT).}
\label{tab:human_eval}
\end{table*}

Except for the RoLLaMA model, which has been specifically fine-tuned for Romanian language tasks, all the other models support Romanian only incidentally.

Similarly to NLLB, we have the same objective of translating between Aromanian, Romanian, and English, so we train the LLMs with samples from all six possible directions. 
In the case of the four models, fine-tuning for more than $1$ epoch leads to an increased loss on the \textit{dev} split. Thus, we keep only the checkpoints fine-tuned for $1$ epoch with sample packing \cite{krell2021efficient}. 
An example of a prompt for the LLaMA 3.1 model can be found in \autoref{tab:prompt}. More details about each model's training hyper-parameters and prompt format are available in \autoref{appendixE}.

\section{Evaluation}
\label{sec:eval}
\subsection{Automatic Evaluation}
\label{sec:aEval}

We evaluate all the models in all possible directions between Aromanian, Romanian, and English. We note here that the English references are machine-translated and the results involving English should be taken with reservation. However, we provide the evaluation scores here for completeness.
The automatic metrics that we employ are BLEU \cite{post-2018-call}, ChrF++ \cite{popovic2015chrf}, and BERTScore \cite{zhang2019bertscore}. We report the ChrF++ scores for all the models in \autoref{chrfres_paper}. All other results are available in \autoref{appendixEVALS} and \autoref{appendixEVALS_dev_set}

Overall, for English-Romanian pairs, GPT-4o obtains the strongest scores in both directions, even when translating from noisy synthetic English into Romanian. Chances that the model has been previously exposed to the out-of-domain test set are very small, since the \textit{Writings} are not available online. Given the synthetic nature of the English data, we cannot draw any conclusions with respect to translation quality into English.

More importantly, the scores for the Aromanian-Romanian language pairs obtained by any of the trained models are considerably higher than those obtained by GPT-4o. Among the trained models, the NLLB family consistently obtains higher scores than their fine-tuned instruction LLM counterparts, regardless of the automatic metric (see \autoref{chrfres_paper} and  \autoref{appendixEVALS}).

When translating into Aromanian ($\rightarrow$rup), the differences are noticeable – in every single case NLLB-type models perform better than LLM-based translation models. Still, when translating into a well-resourced language such as English or Romanian, the performance gap narrows, with the models differing by only a few points. In-domain evaluation on the \textit{dev} set (in \autoref{appendixEVALS_dev_set}) shows significantly higher scores than out-of-domain evaluation on \textit{test} set, with $+14$ points in ChrF and double the BLEU scores. On the \textit{dev} set, the differences between NLLB models and LLMs are less pronounced.

The evaluation scores do not indicate that Aromanian translation benefits more from using RoLLaMA, a model trained on Romanian instructions, or TowerInstruct v0.2, a model designed specifically for MT tasks.

All types of models struggle to generate high-quality Aromanian output, as evidenced by the overall scores. The differences range from 5 to 15 points, with translations into Aromanian scoring lower than translations from Aromanian. \autoref{appendixEVALS} provides additional metrics that align with our findings using ChrF++. We note that \citet{costa2022no} found that a $+1.0$ increase in ChrF++ is almost always noticeable by human evaluators.

Manual inspection of translations reveals that the models tend to translate sentences word-for-word, which misses word overlaps in BLEU-based metrics for out-of-domain evaluation on mixed-genre literary texts, news, and essays. More details on manual evaluation are presented in the next section.



\subsection{Human Evaluation}
Two human evaluators participated in this study. Both are native Aromanian speakers from Romania, fluent in both Romanian and Aromanian. They self-identify as speakers of the Aromanian Grămustean and Cipan dialects, classified as Type A in the typology of \citet{caragiuDictionar} and \citet{pascaru_kahl_rramanj_2017}, which is also the main Aromanian variety in the dataset.

We do not evaluate for English here. 
The evaluated sentences contain human and machine translations in equal proportions, and therefore each annotator evaluates a total of $80$ samples. For MT, we use our NLLB 600M fine-tuned model.

Annotators are instructed to assign direct assessment scores \cite{graham2013continuous} from 1 to 10 to evaluate the quality of each translation (1 being the lowest score, 10 the highest). This scale is culturally motivated by the fact that it is familiar to the study participants because it is used in Romanian schools at all levels of post-primary school.

They assign a score taking into account three categories: fluency, style, and meaning (logical sense). Fluency refers to how grammatically correct the sentence is. Style denotes how likely the speaker is to build the sentence they are evaluating in the exact same way. Meaning alludes to the logical sense of the sentence.

The evaluation was carried out bilingually for all language pairs. The results are in \autoref{tab:human_eval}, showing that machine translation is rated slightly better than human translation. For each low grade, we ask annotators to provide comments with a description of errors. Lexical errors predominate: annotators identify words that do not exist or are not common in their language variety in both human and machine translated texts, e.g., usage of \textit{bus} - instead of \textit{aftuchină} (En. \textit{bus}); usage of \textit{sala di conțertu} (En. \textit{concert hall}); \textit{inițiativă} (En. \textit{initiative}) are not recognized as valid Aromanian. 

Evaluating machine translation systems for nonstandard language varieties presents several challenges regarding representativeness and meaningfulness: 1) annotators may produce inconsistent results, as they may not be familiar with the full range of Aromanian varieties \cite{pascaru_kahl_rramanj_2017}; 2) machine translation systems generate neologisms derived from Romanian, even when established Aromanian terms exist, particularly when translating contemporary texts or news content; 3) human translations may receive lower ratings from speakers due to dialectal differences.

Therefore, the results of human evaluation for fine-grained quality assessment remain inconclusive. Nonetheless, the fact that $80\%$ of machine translations into and from Aromanian received a perfect score from human annotators suggests that the MT system has certain strengths. 


\section{Online Translation System}
\label{sec:deployment}
\begin{figure}[t]
  \includegraphics[width=\columnwidth]{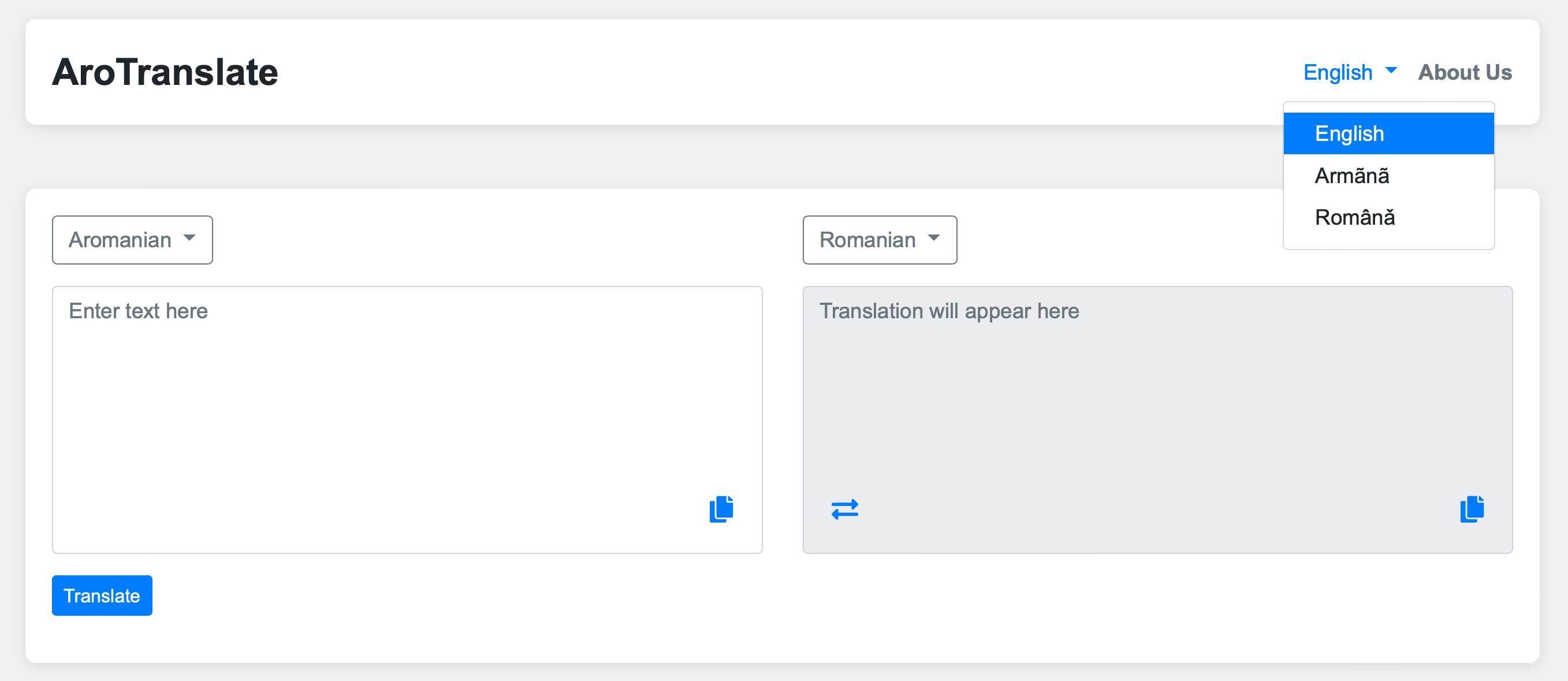}
  \caption{AroTranslate GUI.}
  \label{fig:gui}
\end{figure}

To make the machine translation model accessible to the public and raise awareness of the endangered status of the Aromanian language, we have deployed the system online. The user interface is designed to be lightweight and includes accessibility features such as copying text content and switching between the source and target languages (\autoref{fig:gui}).

\begin{table*}[h]
\centering
\begin{tabular}{l|ll|ll}
\hline
 & \multicolumn{2}{l}{ron$\rightarrow$rup} & \multicolumn{2}{l}{rup$\rightarrow$ron} \\ 
Metric   & \multicolumn{1}{l}{BLEU}  & ChrF++ & \multicolumn{1}{l}{BLEU}  & ChrF++ \\ \hline
FP32 & \multicolumn{1}{l}{17.3}  & 45.03  & \multicolumn{1}{l}{30.95} & 51.01  \\ 
INT8 quantization   & \multicolumn{1}{l}{17.01} & 44.84  & \multicolumn{1}{l}{30.21} & 50.16  \\ 
\end{tabular}
\caption{Differences between the quantized (INT8) and floating point 32 (FP32) versions of the fine-tuned NLLB 600M model, measured on the \textit{test} split.}
\label{fig:quant}
\end{table*}

\subsection{Quantization}

The fine-tuned NLLB-600M model demonstrates the highest performance and has the smallest number of parameters, making it suitable for deployment on a CPU. We quantize the model using the ctranslate2 \cite{klein2020opennmt} engine to 8-bit integer weights. Furthermore, the engine applies additional optimizations, including layer fusion, padding removal, batch reordering, and more. 

Regarding the performance loss typically associated with quantization, we observe minimal differences in the automatic metrics used. As shown in \autoref{fig:quant}, which analyzes the Aromanian-Romanian direction on the \textit{test} split, the differences are measured at less than $1$ BLEU point. Since the models are fine-tuned for sentence-level translations, the system splits user input into sentences and processes them in parallel. The system achieves a translation speed of approximately $65$ tokens per second.

\subsection{Orthographic Converter}
\label{sec:ortoConv}
The system generates translations using the \textit{Cunia} orthography for Aromanian. Additionally, we also implement the option to convert the output to the linguistically-motivated \textit{DIARO} standard. The only ambiguous letter in Cunia standard is <ã> because it can represent either the close central vowel  \textipa{[1]} (e.g. cãndu / when /  \textipa{/k1n"dw/}) or the mid central vowel \textipa{[@]} (e.g., \textipa{[@]}  / tricurã / passed by \textipa{/tri"ku"r@/}; ãncã / yet / \textipa{/1n"k@/}).

For this special case, we employ a two-fold solution: (1) word normalization, where the word is normalized and then matched with the most frequent replacement for that specific word form from a dictionary, and (2) a character 4-gram based solution where for each <ã> in each word we analyse the two neighbouring letters to the right and left and construct frequency masks for these 4-grams. During inference, after replacing all other diacritics, the system selects each 4-gram and chooses the most frequent of its two possible replacements. If neither approach succeeds, the default replacement is biased towards the mid central vowel  \textipa{[@]} based on phonological suggestions from native speakers. The accuracy of this solution is roughly $96\%$.

\section{Conclusions}
Building a dialectal machine translation system for a low-resource language comes with several challenges – from the lack of standardization and small online presence of the language to the diverse varieties specific to different countries. 

Previous experiments show that modern neural machine translation systems are unable to learn from extremely small data sets of 3,000 pair of sentences, even for languages that are part of the same dialectal continuum such as Romanian and Aromanian \cite{petrariu2024multilingual}. In our work, we provide the steps to construct a parallel corpus for Aromanian, which further requires training a multilingual sentence embedder. We show that a diverse parallel dataset of $79$k examples is sufficient for fine-tuning pre-trained transformer models to do low-resource and dialectal machine translation. 
The preexisting knowledge of Romanian in both LLM transformers and sequence-to-sequence models enhances generalization and knowledge transfer capabilities. However, it also introduces bias and raises concerns about representativeness, as the generated Aromanian shows strong Romanian influence and similarities.

The introduction of English synthetic data in the training phase proves useful to extend the translation directions beyond Romanian. However, future work would require measuring the impact of translation errors in synthetic data.

Based on automatic evaluation, the smaller seq-to-seq models are better translators than LLM-based English-centric or multilingual fine-tuned models (that have not been pre-trained on Romanian). NLLB-based models outperform the LLM-based pre-trained models in terms of automatic metrics and have the advantage of being light-weight and easy to deploy on CPU using int8 quantization.

Regarding human evaluation, due to the relatively small size of the community and lack of access to a large pool of native speakers, we had difficulties conducting an extensive human evaluation. Furthermore, we identify several difficulties in conducting human evaluations on low-resource, unstandardized dialectal varieties, and future studies are mandatory to quantify the limitations of our MT system geographically and across language varieties.

With this work, we hope to bring Aromanian into the spotlight of machine translation and corpus-based linguistic research and contribute to preservation efforts for this endangered language.

\section{Limitations}
Aromanian is a very low-resource language and is not standardized, having several varieties. As a result, translation quality is experimental and often times sub-par to what a user might expect from a translation system for medium- to high-resource languages. This model is biased to favour some Aromanian varieties more than others. Translation errors may occur when certain information is missing or altered. There is also concern about the model's bias, as the Bible is a dominant source in our corpus. 
Last but not least, this work already mentions a limitation in terms of lack of availability to carry an in-depth speaker-centric human evaluation of the MT system. This is an issue that we are looking into in future work. 
We release the corpus under the Creative Commons Attribution-NonCommercial 4.0 International License \footnote{\url{https://creativecommons.org/licenses/by-nc/4.0/}}.

\section{Ethical Considerations}
We were able to collect and compile parallel translated texts from multiple sources. Note that for each text source we have received permission to use the text, verified appropriate licensing agreements, or confirmed fair-use applicability for academic research purposes.
The Google Translate API calls cost around $180$\$ in credits and the OpenAI API calls cost around $10$\$. Model training was performed on cloud-based servers (A100 and RTX 4090 GPUs) rented by the hour, incurring a total cost of approximately $500$\$.

\section*{Acknowledgments}
This work would not have been possible without the support of the Aromanian Community from Romania (CARo), who helped us with the data collection processes, system testing, and gave us invaluable advice. We specifically acknowledge Florentina Costea, Cristian Jiru, and Nicolae Todică.
Regarding data collection, we are grateful to Kira Mantsu and George Vrana for all their help. The human evaluation process would not have been possible without the generous help of Georgiana Stavrositu, Sebastian Florea, and Zoia Dragomir.
The machine translation system is hosted with the generous support of the University of Bucharest \url{https://arotranslate.unibuc.ro}.

This research is partially supported by the project ``Romanian Hub for Artificial Intelligence - HRIA'', Smart Growth, Digitization and Financial Instruments Program, 2021-2027, MySMIS no. 334906 and by InstRead: Research Instruments for the Text Complexity, Simplification and Readability Assessment  CNCS - UEFISCDI project number PN-IV-P2-2.1-TE-2023-2007.

\clearpage

\bibliography{coling_v1.bib}

\clearpage

\onecolumn
\appendix
\section{Aromanian-Romanian Sample Pairs}
\label{sec:appendixA}
\begin{table*}[h]
  \centering
  \small
  \resizebox{\linewidth}{!}{
  \begin{tabular}{p{0.08\textwidth}p{0.42\textwidth}p{0.42\textwidth}}
    \hline
    \textbf{Source} & \textbf{Aromanian} & \textbf{Romanian} \\
    \hline
    Bible & 
      Și vasiljelu Ptolometi muri după trei dzali, shi-ashchirladzlji cari eara pit tsitãts lj-vatamara oaminjlji di-a loclui. \newline Ti-agiutarea-a tsilor far di mindueari-arauda unã giudicata cu itsralji, a tinirlui om ti-unã cãnushteari shi bunã mindueari. &
      Și regele Ptolomeu a murit a treia zi, și ostașii care erau în cetăți au fost uciși de către locuitori. \newline Pentru a prilejui celor fără gând rău o judecată isteață, omului tânăr cunoștință și bună cugetare. \\
    \hline
    Lyrics T. &
      Pisti giuguri trecu anarga, lãi niori \newline Nicuchirã cu curunã: &
      Peste juguri trec încet, nori negri \newline Ar trimite-o el acasă, \\
    \hline
    MPC-rup &
      Unã caprã mushcrã shi unã ghesucãnutã pãshteau di unã parte, shi chipurle lã asunau: cing-cing. \newline Tsachilji giuca deavãrliga, arsãrea - tuts aleptsã, musheats, cu lãna lor ca mãtasea, di-lji yinea shi-al Tegã se-arsarã shi s-lu stringã-n bratsã picurarlu shi s-lu bashe, ahtare harauã lu-avea loatã! &
      O capră cu botul bălțat și alta roșcată - pășteau răzlețite și tot le sunau tălăngile de la gât: cing-cing-cling. \newline Țapii se zbenguiau prin jur, țupăiau - frumoși toți, dragii, cu părul lor mătăsos, de-i venea lui Tegă, de bucurie, să-l ia în brațe pe cioban, să-l strângă la piept și să-l sărute, nestăpânit! \\
    \hline
    Phrases &
      nu shtiu si mpartu palji la doi yumari \newline misuru stealili &
      fi prost \newline hoinări \\
    \hline
    Avdhela &
      Di toamnã-ascuturatã Ma, \newline s-ti caftu nji-easti fricã, &
      Se scutură de toamnă, \newline Să te cer îmi este teamă, \\
    \hline
    Radio &
      Deputatlu USR Cristian Seidler spusi cã proiectul di nom ari un impact multu modestu ti hãrgili bugetare cu pensiile spetsiali. 
      &
      Deputatul USR Cristian Seidler a afirmat că proiectul de lege are un impact foarte modest asupra cheltuielilor bugetare cu pensiile speciale.  \\
    \hline
    Aromanian \newline Poetry &
      Am un alt livend, di-Avdelǎ, alǎvdatlu-atsel di NUShI, te da njeate cãt sh-la-aushi; \newline Nã shcãmbã fãrã suflit i semn di tine, tutǎ shi salta siminatã di tine, tu ubor... &
      În Avdela am un june tULLIU NUȘI - vorbă cerească versu-n veci o să-i trăiască ! \newline O piatră doar aminte de tine mai aduce, și-o salcie sădită de mâna ta, cândva... \\
    \hline
    Tom \newline Sawyer &
      Arucã funea! \newline Mutri anvãrliga, canã... tu niheam di oarã u avea cartea tru mãnã. &
      Dă și parâma! \newline Aruncă o privire de jur-împrejur, nu era nimeni... în clipa următoare, ținea cartea în mâini. \\
    \hline
    Divine \newline Comedy &
      Shi-agãrsheashti-n frixea tsi-lji da dãgoarea \newline tu doilji sherchi-mpriunats trupeashti &
      pârjolul să-și ia ceva, căci la copil ia seamă \newline în cei doi șerpi împreunați \\
    \hline
    Modern \newline poetry &
      mashi suflitlu nu ari moarti \newline aclo iu suntu farurli apãryisiti. &
      doar sufletul este fără moarte \newline acolo unde sunt farurile abandonate. \\
    \hline
    Tao \newline Te Ching &
      Pots s-ti tradz ditu minduirea a ta \newline Imnji dupã ea sh-nu-ari bitisitã. &
      Poți să te dai înapoi din propria ta minte \newline urmează-l și nu are sfârșit. \\
    \hline
    The Little \newline Prince &
      Gioaca cor cathi gioi cu featili di-n hoara. \newline Elj avdu daima mashi alavdarli. &
      Se duc să joace, joia, cu fetele din sat. \newline Vanitoșii nu aud niciodată decât laudele. \\
    \hline
    Writings &
      Li-alidzea ayonjia, unã dupu altã, ashi cum beai unã scafã cu apã! \newline  Bãnarã deadunu tu Vãryãrie. &
      Le citea repede, una după alta, așa cum ai bea un pahar cu apă! \newline Trăiseră împreună în Bulgaria. \\
    \hline
  \end{tabular}%
  }
  \caption{Aromanian-Romanian sample pairs from various sources. The texts have been converted in the \textit{Cunia} standard.}
  \label{citation-guide}
\end{table*}

\begin{table*}[h]
  \centering
  \small
  \resizebox{\linewidth}{!}{
  \begin{tabular}{p{0.08\textwidth}p{0.42\textwidth}p{0.42\textwidth}}
    \hline
    \textbf{Source} & \textbf{Aromanian} & \textbf{Romanian} \\
    \hline
    Bible & 
      Și vasiľelu Ptolometi muri dupâ trei dzali, și-așchirladzľi cari eara pit țităț ľ-vatamara oamińľi di -a loclui.\newline Ti-agiutarea-a ților far di mindueari-arauda unâ giudicata cu ițraľi, a tinirlui om ti-unâ cânușteari și bunâ mindueari. &
      Și regele Ptolomeu a murit a treia zi, și ostașii care erau în cetăți au fost uciși de către locuitori. \newline Pentru a prilejui celor fără gând rău o judecată isteață, omului tânăr cunoștință și bună cugetare. \\
    \hline
    Lyrics T. &
      Pisti giuguri trecu anarga, lâi niori \newline nicuchirâ cu curunâ: &
      Peste juguri trec încet, nori negri \newline Ar trimite-o el acasă, \\
    \hline
    MPC-rup &
      Unâ caprâ mușcră și unâ ghesucănută pășteau di unâ parte, și chipurle lâ asunau: cing-cing.\newline Țachiľi giuca deavârliga, arsărea-tuț alepțâ, mușeaț, cu lâna lor ca mătasea, di-ľi yinea și-al Tegâ se-arsară și s-lu stringă-n brațâ picurarlu și s-lu bașe, ahtare harauâ lu-avea loatâ! &
      O capră cu botul bălțat și alta roșcată - pășteau răzlețite și tot le sunau tălăngile de la gât: cing-cing-cling. \newline Țapii se zbenguiau prin jur, țupăiau - frumoși toți, dragii, cu părul lor mătăsos, de-i venea lui Tegă, de bucurie, să-l ia în brațe pe cioban, să-l strângă la piept și să-l sărute, nestăpânit! \\
    \hline
    Phrases &
      nu știu si mpartu paľi la doi yumari \newline misuru stealili &
      fi prost \newline hoinări \\
    \hline
    Avdhela &
      Di toamnă-ascuturată Ma, \newline s-ti caftu ńi-easti fricâ, &
      Se scutură de toamnă, \newline Să te cer îmi este teamă, \\
    \hline
    Radio &
      Deputatlu USR Cristian Seidler spusi că proiectul di nom ari un impact multu modestu ti hârgili bugetare cu pensiile spețiali.  &
      Deputatul USR Cristian Seidler a afirmat că proiectul de lege are un impact foarte modest asupra cheltuielilor bugetare cu pensiile speciale. \\
    \hline
    Aromanian \newline Poetry &
      Am un alt livend, di-Avdelă, alâvdatlu-ațel di NUȘI, te da ńeate cât ș-la-auși; \newline nâ șcâmbâ fârâ suflit i semn di tine, tutâ și salta siminată di tine, tu ubor... &
      În Avdela am un june tULLIU NUȘI - vorbă cerească versu-n veci o să-i trăiască ! \newline O piatră doar aminte de tine mai aduce, și-o salcie sădită de mâna ta, cândva... \\
    \hline
    Tom \newline Sawyer &
      Arucâ funea! \newline Mutri anvârliga, canâ... tu niheam di oarâ u avea cartea tru mânâ. &
      Dă și parâma! \newline Aruncă o privire de jur-împrejur, nu era nimeni... în clipa următoare, ținea cartea în mâini. \\
    \hline
    Divine \newline Comedy &
      Și-agârșeaști-n frixea ți-ľi da dăgoarea \newline tu doiľi șerchi-mpriunaț trupeaști &
      pârjolul să-și ia ceva, căci la copil ia seamă \newline în cei doi șerpi împreunați \\
    \hline
    Modern \newline poetry &
      mași suflitlu nu ari moarti \newline aclo iu suntu farurli apâryisiti. &
      doar sufletul este fără moarte \newline acolo unde sunt farurile abandonate. \\
    \hline
    Tao \newline Te Ching &
      Poț s-ti tradz ditu minduirea a ta \newline Imńi dupâ ea ș-nu-ari bitisitâ. &
      Poți să te dai înapoi din propria ta minte \newline urmează-l și nu are sfârșit. \\
    \hline
    The Little \newline Prince &
      Gioaca cor cathi gioi cu featili di-n hoara. \newline Eľ avdu daima mași alavdarli. &
      Se duc să joace, joia, cu fetele din sat. \newline Vanitoșii nu aud niciodată decât laudele. \\
    \hline
    Writings &
      Li-alidzea ayońia, unâ dupu altâ, ași cum beai unâ scafâ cu apâ! \newline  Bânarâ deadunu tu vâryârie. &
      Le citea repede, una după alta, așa cum ai bea un pahar cu apă! \newline Trăiseră împreună în Bulgaria. \\
    \hline
  \end{tabular} %
  }
  \caption{Aromanian-Romanian sample pairs from various sources, converted to an approximate form of the \textit{DIARO} standard.}
  \label{citation-guide}
\end{table*}

Our final data release also contains $27.3$k dictionary pairs that we do not use to train the models. When training our initial machine translation models, we did not observe any improvements when using word-aligned dictionary entries. The entries are extracted from Papahagi's Aromanian dictionary \cite{dictionarPapahagi}.

\clearpage

\section{Corpus Statistics}
\label{sec:appendixB}
\begin{table*}[!htbp]
\centering
\begin{tabular}{llrrrr}
\hline
language & source & words & unique words &  words/sentence & type-token ratio \\
\hline
\multirow{12}{*}{Aromanian} 
& Bible & 493018 & 69780 & 16.15 & 0.14 \\
& Lyrics T. & 16039 & 7586 & 3.98 & 0.47 \\
& MPC-rup & 508090 & 74419 & 27.1 & 0.14 \\
& Phrases & 40941 & 9972 & 4.78 & 0.24 \\
& Avdhela & 7839 & 3567 & 3.48 & 0.45 \\
& Radio & 41036 & 12708 & 12.15 & 0.30 \\
& Aromanian Poetry & 36649 & 10433 & 11.69 & 0.28 \\
& Tom Sawyer & 23177 & 10325 & 13.13 & 0.44 \\
& Divine Comedy & 1269 & 747 & 4.83 & 0.58 \\
& Modern poetry & 11897 & 6336 & 5.3 & 0.53 \\
& Tao Te Ching & 5653 & 1886 & 2.62 & 0.33 \\
& The little prince & 3619 & 1625 & 5.81 & 0.44 \\
& Writings Collection & 17893 & 6646 & 12.81 & 0.37 \\
\hline
\multirow{12}{*}{Romanian} 
& Bible & 580288 & 42487 & 19.0 & 0.07 \\
& Lyrics T. & 16923 & 7022 & 4.2 & 0.41 \\
& MPC-rup & 494098 & 53549 & 26.36 & 0.10 \\
& Phrases & 43590 & 7484 & 5.08 & 0.17 \\
& Avdhela & 8186 & 3330 & 3.63 & 0.40 \\
& Radio & 47588 & 12291 & 14.09 & 0.25 \\
& Aromanian Poetry & 38719 & 11126 & 12.35 & 0.28 \\
& Tom Sawyer & 24665 & 9296 & 13.97 & 0.37 \\
& Divine Comedy & 1397 & 744 & 5.31 & 0.53 \\
& Modern poetry & 14146 & 6412 & 6.31 & 0.45 \\
& Tao Te Ching & 4877 & 1740 & 2.26 & 0.35 \\
& The little prince & 3977 & 1717 & 6.38 & 0.43 \\
& Writings Collection & 19328 & 6398 & 13.84 & 0.33 \\
\hline
\multirow{12}{*}{English} 
& Bible & 622430 & 28266 & 20.38 & 0.04 \\
& Lyrics T. & 20277 & 5067 & 5.03 & 0.24 \\
& MPC-rup & 518118 & 34576 & 27.64 & 0.06 \\
& Phrases & 49025 & 4950 & 5.72 & 0.10 \\
& Avdhela & 9466 & 2671 & 4.2 & 0.28 \\
& Radio & 53286 & 8348 & 15.78 & 0.15 \\
& Aromanian Poetry & 41069 & 7692 & 13.1 & 0.18 \\
& Tom Sawyer & 29411 & 6491 & 16.67 & 0.22 \\
& Divine Comedy & 1457 & 675 & 5.54 & 0.46 \\
& Modern poetry & 16959 & 4892 & 7.56 & 0.28 \\
& Tao Te Ching & 5089 & 1597 & 2.36 & 0.31 \\
& The little prince & 4258 & 1380 & 6.83 & 0.32 \\
& Writings Collection & 20998 & 4795 & 15.03 & 0.22 \\
\hline
\end{tabular}
\caption{\label{citation-guide}
       Corpus statistics for each source, computed on the version converted to \textit{Cunia} standard. 
}
\end{table*}
\clearpage

\section{LaBSE Training Hyper-paramateres}
\label{sec:appendixLABSE}
\begin{table*}[h]
  \centering
  \begin{tabular}{cc}
    \hline
    Training steps & 150000  \\
    Batch size & 8  \\
    Margin & 0.3 \\
    Optimizer & Adafactor \cite{shazeer2018adafactor} \\
    Learning rate &  $1\text{e-}5$ \\
    Clip threshold &  $1.0$ \\
    \hline
    \end{tabular}

  \caption{\label{citation-guide}
       LaBSE training hyperparameters, Appendix C.
  }
\end{table*}

\section{NLLB Training Hyper-parameters}
\label{appendixD}
\begin{table*}[h]
  \centering
  \begin{tabular}{cc}
    \hline
    Training steps & 100000  \\
    Batch size & 8  \\
    Optimizer & Adafactor \cite{shazeer2018adafactor} \\
    Scheduler & constant  \\
    Learning rate &  $1\text{e-}4$ \\
    Weight Decay &  $1\text{e-}3$ \\
    Clip threshold &  $1.0$ \\
    Maximum Sequence Length & 512 \\
    \hline
    \end{tabular}

  \caption{\label{citation-guide}
       NLLB training hyperparameters for both the 600M and 1.3B distilled versions.
  }
\end{table*}

\section{LLM Training}
\label{appendixE}

\subsection{Hyper-parameters}
\begin{table*}[h]
  \centering
  \begin{tabular}{cc}
    \hline
    Effective batch size & 32  \\
    Number of epochs & 4 \\
    Learning rate & $2\text{e-}5$  \\
    LR scheduler &  cosine \\
    Warmup steps & 100 \\
    Weight decay &  $0.001$ \\
    Adam $\beta_{1}$ &  $0.9$ \\
    Adam $\beta_{2}$ &  $0.999$ \\
    Adam $\epsilon$ & $1\text{e-}8$ \\
    Maximum Sequence Length & 512 \\
    \hline
    \end{tabular}

  \caption{\label{citation-guide}
       Training hyper-parameters for all the fine-tuned LLMs.
  }
\end{table*}

\subsection{Prompt Format}
We format the prompts of the LLaMA 3.1 8B Instruct and RoLLaMA 3 8B Instruct models using the standard LLaMA 3.1 Instruct prompt format \cite{dubey2024llama}. 
For the Qwen and TowerInstruct models, we use their specific instruction prompt template \cite{yang2024qwen2, alves2024tower}.
\begin{table*}[h]
  \centering
  \begin{tabular}{lp{12cm}}
    \hline
    System & 
    \texttt{<|start\_header\_id|>system<|end\_header\_id|>}\newline \newline You are a helpful assistant capable of accurately translating between Aromanian, Romanian, and English. \newline \newline \texttt{<|eot\_id|>}  \\
    User & \texttt{<|start\_header\_id|>user<|end\_header\_id|>}\newline \newline Translate this sentence from Romanian to Aromanian\newline Adu-ți aminte că moartea nu zăbovește și hotărârea morții nu ți s-a arătat.\texttt{<|eot\_id|>} \\
    Model & \texttt{<|start\_header\_id|>assistant<|end\_header\_id|>}\newline \newline Adu-ts aminti cã moartea nu-amãnã shi-apofasea-a moartiljei nu tsã si-ari spusã. \\
    \hline
    \end{tabular}

  \caption{\label{citation-guide}
       Prompt format for LLaMA 3.1 8B Instruct and RoLLaMA 3 8B Instruct. Note that this is an example, and the source and target language are not fixed, and do include \textit{English} as well.
  }
\end{table*}

\begin{table*}[h]
  \centering
  \begin{tabular}{lp{12cm}}
    \hline
    System & 
    \texttt{<|im\_start|>system}\newline You are a helpful assistant capable of accurately translating between Aromanian, Romanian, and English.\texttt{<|im\_end|>}  \\
    User & \texttt{<|im\_start|>user}\newline Translate this sentence from Romanian to Aromanian\newline Adu-ți aminte că moartea nu zăbovește și hotărârea morții nu ți s-a arătat.\texttt{<|im\_end|>} \\
    Model & \texttt{<|im\_start|>assistant}\newline Adu-ts aminti cã moartea nu-amãnã shi-apofasea-a moartiljei nu tsã si-ari spusã.\texttt{<|im\_end|>} \\
    \hline
    \end{tabular}

  \caption{\label{citation-guide}
       Prompt format for Qwen2 7B Instruct. Note that this is an example, and the source and target language are not fixed, and do include \textit{English} as well.
  }
\end{table*}

\begin{table*}[h]
  \centering
  \begin{tabular}{lp{12cm}}
    \hline
    User & \texttt{<|im\_start|>user}\newline Translate this sentence from Romanian to Aromanian\newline Adu-ți aminte că moartea nu zăbovește și hotărârea morții nu ți s-a arătat.\texttt{<|im\_end|>} \\
    Model & \texttt{<|im\_start|>assistant}\newline Adu-ts aminti cã moartea nu-amãnã shi-apofasea-a moartiljei nu tsã si-ari spusã.\texttt{<|im\_end|>} \\
    \hline
    \end{tabular}

  \caption{\label{citation-guide}
       Prompt format for TowerInstruct 7B v0.2. Note that this is an example, and the source and target language are not fixed, and do include \textit{English} as well.
  }
\end{table*}
\clearpage

\section{Automatic Evaluation Results - \textit{test} Set}
\label{appendixEVALS}
\begin{table*}[h]
\centering
\begin{tabular}{lrr|rr|rr}
\hline
model & ron$\rightarrow$rup & rup$\rightarrow$ron & eng$\rightarrow$rup & rup$\rightarrow$eng & eng$\rightarrow$ron & ron$\rightarrow$eng \\
\hline
GPT-4o & 3.5 & 19.4 & 2.9 & 19.0 & 46.5 & 66.7 \\
Qwen2 7B Instruct     & 15.2                  & 28.8                  & 10.4                  & 32.4                  & 35.4                  & 63.0                  \\
TowerInstruct v0.2    & 16.3                  & 29.5                  & 7.3                   & 23.7                  & 27.1                  & 62.1                  \\
LLaMA 3.1 8B Instruct & 15.9                  & 30.3                  & 10.8                  & 33.7                  & 38.6                  & 65.1                  \\
RoLLaMA               & 14.8                  & 29.7                  & 11.1                  & 32.8                  & 38.1                  & 64.4                  \\ 
\hline
NLLB 1.3B             & 16.3                  & 31.9                  & 12.8                  & 35.7                  & 41.4                  & 63.0                  \\
NLLB 600M             & 17.0                  & 30.9                  & 13.5                  & 35.0                  & 39.1                  & 62.2 \\
\hline
\end{tabular}
\caption{
BLEU evaluation scores on the \textit{test} set align with the ChrF++ results from Table 4. NLLB models consistently outperform LLMs for Aromanian translations in both directions, with a notable discrepancy between translations into and from Aromanian.
}
\label{bleures}
\end{table*}


\begin{table*}[h]
\centering
\begin{tabular}{lcc|cc|cc}
\hline
model & ron$\rightarrow$rup & rup$\rightarrow$ron & eng$\rightarrow$rup & rup$\rightarrow$eng & eng$\rightarrow$ron & ron$\rightarrow$eng \\
\hline
GPT-4o & 0.866 & 0.816 & 0.857 & 0.829 & 0.968 & 0.980 \\
Qwen2 7B Instruct & 0.905 & 0.896 & 0.880 & 0.903 & 0.955 & 0.976 \\
TowerInstruct v0.2 & 0.904 & 0.894 & 0.840 & 0.870 & 0.923 & 0.973 \\
LLaMA 3.1 8B Instruct & 0.908 & 0.899 & 0.888 & 0.907 & 0.959 & 0.978 \\
RoLLaMA & 0.908 & 0.898 & 0.887 & 0.904 & 0.957 & 0.978 \\
\hline
NLLB 1.3B & 0.913 & 0.905 & 0.904 & 0.913 & 0.964 & 0.977 \\
NLLB 600M & 0.914 & 0.905 & 0.904 & 0.913 & 0.964 & 0.976 \\
\hline
\end{tabular}
\caption{
BERTScore evaluation on the \textit{test} set. }
\label{bertres}
\end{table*}

\clearpage
\section{Automatic Evaluation Results - \textit{dev} Set}
\label{appendixEVALS_dev_set}
\begin{table*}[h]
\centering
\begin{tabular}{lrr|rr|rr}
\hline
model                 & ron$\rightarrow$rup & rup$\rightarrow$ron & eng$\rightarrow$rup & rup$\rightarrow$eng & eng$\rightarrow$ron & ron$\rightarrow$eng \\
Qwen2 7B Instruct     & 31.9    & 50.7    & 23.9    & 51.7    & 49.8    & 67.8    \\
TowerInstruct v0.2    & 34.5    & 52.6    & 26.4    & 53.1    & 52.1    & 70.5    \\
LLaMA 3.1 8B Instruct & 34.8    & 53.6    & 26.6    & 52.4    & 52.4    & 69.5    \\
RoLLaMA               & 34.4    & 53.0    & 26.3    & 51.4    & 52.4    & 68.9    \\
\hline
NLLB 1.3B             & 33.7    & 53.2    & 26.2    & 52.7    & 52.1    & 68.5    \\
NLLB 600M             & 33.6    & 52.0    & 25.6    & 52.2    & 50.5    & 67.3    \\  
\hline
\end{tabular}
\caption{
BLEU evaluation scores on the \textit{dev} set.
}
\label{bleures}
\end{table*}

\begin{table*}[h]
\centering
\begin{tabular}{lrr|rr|rr}
\hline
model & ron$\rightarrow$rup & rup$\rightarrow$ron & eng$\rightarrow$rup & rup$\rightarrow$eng & eng$\rightarrow$ron & ron$\rightarrow$eng \\
\hline
Qwen2 7B Instruct     & 61.9                  & 68.1                  & 55.6                  & 67.6                  & 70.1                  & 80.3                  \\
TowerInstruct v0.2    & 63.3                  & 69.1                  & 56.9                  & 68.8                  & 71.9                  & 82.1                  \\
LLaMA 3.1 8B Instruct & 63.7                  & 69.8                  & 57.2                  & 68.5                  & 72.0                  & 81.6                  \\
RoLLaMA               & 63.7                  & 69.4                  & 57.2                  & 68.1                  & 71.9                  & 81.1                  \\
\hline
NLLB 1.3B             & 63.7                  & 69.9                  & 57.7                  & 68.4                  & 72.2                  & 80.8                  \\
NLLB 600M             & 63.6                  & 69.2                  & 57.4                  & 67.9                  & 71.1                  & 80.0                 \\

\hline
\end{tabular}
\caption{
ChrF++ evaluation scores on the \textit{dev} set.
}
\label{bleures}
\end{table*}

Compared to the out-of-domain \textit{test} set, 
the evaluation scores on the \textit{dev} set are considerably higher and the differences between LLMs and NLLB-type models is smaller.

ChrF signature is \texttt{nrefs:1|case:mixed|eff:yes|nc:6|nw:0|space:no|version:2.4.3} and BLEU signature is \texttt{nrefs:1|case:mixed|eff:no|tok:13a|smooth:exp|version:2.4.3}.

\end{document}